\documentclass{article}

\usepackage[preprint,nonatbib]{neurips_2023}

\usepackage[utf8]{inputenc} % allow utf-8 input
\usepackage[T1]{fontenc}    % use 8-bit T1 fonts
\usepackage{hyperref}       % hyperlinks
\usepackage{url}            % simple URL typesetting
\usepackage{booktabs}       % professional-quality tables
\usepackage{amsfonts}       % blackboard math symbols
\usepackage{nicefrac}       % compact symbols for 1/2, etc.
\usepackage{microtype}      % microtypography
\usepackage{xcolor}         % colors
\usepackage{graphicx}
\usepackage{subfigure}
\usepackage{subcaption}
\usepackage{wrapfig}
\usepackage{amssymb}
\usepackage{amsmath}

\newcommand{\KL}{D_{\mathrm{KL}}}

\title{Towards Cost Sensitive Decision Making}

\author{%
Yang Li\\
Department of Computer Science\\
UNC-Chapel Hill\\
\texttt{yangli95@cs.unc.edu}\\
\And
Junier Oliva\\
Department of Computer Science\\
UNC-Chapel Hill\\
\texttt{joliva@cs.unc.edu}\\
}

\begin{document}

\maketitle

\begin{abstract}
Many real-world situations allow for the acquisition of additional relevant information when making decisions with limited or uncertain data. However, traditional RL approaches either require all features to be acquired beforehand (e.g. in a MDP) or regard part of them as missing data that cannot be acquired (e.g. in a POMDP). In this work, we consider RL models that may actively acquire features from the environment to improve the decision quality and certainty, while automatically balancing the cost of feature acquisition process and the reward of task decision process. We propose the Active-Acquisition POMDP and identify two types of the acquisition process for different application domains. In order to assist the agent in the actively-acquired partially-observed environment and alleviate the exploration-exploitation dilemma, we develop a model-based approach, where a deep generative model is utilized to capture the dependencies of the features and impute the unobserved features. The imputations essentially represent the beliefs of the agent. Equipped with the dynamics model, we develop hierarchical RL algorithms to resolve both types of the AA-POMDPs. Empirical results demonstrate that our approach achieves considerably better performance than existing POMDP-RL solutions.
\end{abstract}

\section{Introduction}
Recently, machine learning models for sequential decision making have made significant progress due to the development of reinforcement learning (RL). These models have achieved remarkable success in many application domains, such as games \cite{mnih2015human,silver2016mastering,silver2017mastering}, robotics control \cite{finn2016guided,levine2016end,polydoros2017survey,haarnoja2018composable,niroui2019deep} and medical diagnosis \cite{ling2017learning,peng2018refuel,coronato2020reinforcement,yu2021reinforcement}. 
However, the current RL paradigm is incongruous with the expectation of many real-world decision-making systems. First, for fully-observed Markov decision processes (MDPs), the features at each decision step are assumed to be fully observed. In situations like medical diagnosis, some features, such as MRI, might be expensive to obtain; some features might even pose a risk to the patient, such as X-Ray. Furthermore, acquiring all features at each step may create redundancy, as some features will not change within the adjacent time frames. Therefore, the intelligent decision-making systems are expected to automatically balance the cost of feature acquisition and the improvement of decision by acquiring only the necessary information. Second, for partially-observed Markov decision processes (POMDPs), the observation at each step is determined by an unknown observation model of the environment, thus no additional information (features) may be obtained to improve the decision.

In stark contrast to the current RL paradigm, human agents routinely reason over instances with incomplete features and decide when and what additional information to obtain. For example, consider clinicians in intensive care units (ICUs), which have to make sequences of treatment decisions for patients at risk. 
Typically, all of the (dynamic) patient information is not known, however, and while knowledge of the patient is critical when deciding what treatment decisions to make, 
due to time/cost/risk constraints the clinician must carefully decide what additional patient attributes (e.g. stemming from a blood sample, or biopsy, etc.) are most worth their cost for better downstream treatment decisions.
%When selecting a treatment, he/she may only measure what is necessary, rather than ordering thorough laboratory test every time, because other measurements may be either predictable from his/her experience or irrelevant to the current decision. In addition, whenever he/she is unsure of the appropriate treatment, he/she can dynamically decide to take more measurements to help determine the decision.
In order to more closely match the needs of many real-world applications, we propose a active acquisition partially observed Markov decision process (AA-POMDP) and develop several novel RL techniques to solve it. Our agent not only makes decisions with incomplete/missing features, but also dynamically determines the most valuable subset of features to obtain at each decision step.

In this work, we identify two types of AA-POMDPs based on how features may be acquired. First, \textbf{Sequential AA-POMDP}, where features are acquired sequentially before a task action is conducted. Here, the later acquisitions will depend on the values of previously acquired features. This type of AA-POMDP is applicable when the feature acquisition actions do not modify the underlying state (such as non-invasive test in medical scenario) and the feature acquisition process takes negligible time compared to the task state changing.
%, so the features acquired early are not obsolete when making the next task decision.
Second, \textbf{Batch AA-POMDP}, where features are acquired simultaneously in a batch. This type of POMDP is suitable for situations where the feature acquisition actions can modify the state or the state changes so quickly relative to acquisition that there is not enough time for a sequential acquisition.

%Due to the dynamic feature acquisition, t
The agent can only observe part of the underlying features when making a decision for the task. Therefore, our model is essentially partially observed and inherits all the difficulties for solving POMDP \cite{monahan1982state,spaan2012partially}. Meanwhile, in contrast to typical POMDP in RL literature, here the observation is controlled by the agent itself, which introduces additional challenges. First, the action space for feature acquisition is exponential to the number of candidate features. That is, for a $d$-dimensional feature space, there are $2^d$ possible acquisition actions in total. The large action space makes it difficult for RL agents to explore efficiently. Second, the feature acquisition process and the task decision process are intimately correlated. The feature acquisition process must collect informative features so that appropriate decisions can be made for the task. Moreover, the task decisions should transit the underlying MDP into appropriate states so that acquisitions can be performed effectively.

In order to deal with the aforementioned challenges, we propose a model-based approach in which a generative model is utilized to capture the dependencies between features (see $\S$~\ref{sec:poseq}). Given a sequence of acquired features and corresponding task actions, the generative model predicts the possible state at the next decision step by imputing the missing features, which represents the beliefs of our agent. The acquisition action and the task action are both determined based on the belief states, which help the agent learn better policy with partial observation. Furthermore, the generative model can assist the agent with an intrinsic reward by assessing the imputation quality of the underlying state, which essentially provides guidance to the acquisition process. In addition, we decompose the acquisition process and the task decision process into a hierarchy, where the high-level task policy takes inputs from the low-level acquisition policy and provides reward signal in return based on its policy uncertainty and value estimates (see $\S$~\ref{sec:csrl} for details).

Our contributions are as follows:
1) We propose the active acquisition partially observed Markov decision process (AA-POMDP), which integrates the feature acquisition process with the task decision process to make decisions taking into account the cost of feature acquisition.
2) We identify two types of the AA-POMDP, Sequential AA-POMDP and Batch AA-POMDP, to accommodate different application requirements.
3) We develop a novel generative model for partially observed sequences that captures the dependencies across features and across time steps. The generative model serves as a surrogate of the task state transition model and assists the agent by estimating the beliefs. We then develop model-based RL agents for both types of the AA-POMDP.
4) We formulate the feature acquisition process and the task decision process into a hierarchical structure and propose hierarchical RL approaches that automatically balance the feature acquisition cost and task reward.
5) We demonstrate the effectiveness of our proposed approach on several benchmark environments and achieve state-of-the-art performance compared to baselines.

\section{Active Acquisition Partially-Observed Markov Decision Process}
A discounted AA-POMDP is an environment defined by a tuple $\mathcal{M}=\left< \mathcal{S}, \mathcal{A}, \mathcal{T}, \mathcal{O}, \mathcal{R}, \mathcal{C}, \gamma \right>$, where $\mathcal{S}$ is the state space and $\mathcal{A} = \mathcal{A}_{f} \cup \mathcal{A}_{c}$ is the joint action space of feature acquisition actions $\mathcal{A}_{f}$ and task control actions $\mathcal{A}_{c}$. $\mathcal{T}: \mathcal{S} \times \mathcal{A} \rightarrow \mathcal{S}$ represents the state transition kernel, which can be deterministic or stochastic. As in ordinary POMDP, the observation space $\mathcal{O}$ is related to $\mathcal{S}$ by the observation/emission model $p(o \mid s', a)$, which defines the probability of observing $o$ when the agent takes action $a$ resulting in a state $s'$. In AA-POMDP, however, the observation model is specified as $p(o \mid s', a_{f})$. I.e., the observation is controlled only by the feature acquisition action $a_{f}$. For a state $s'$ with underlying $d$-dimensional measurable features $x'$ that are unknown beforehand, the feature acquisition action $a_{f}$ will result in acquiring a subset of features $x'_v, v \subseteq \{1,\ldots,d\}$, where $v$ is decoded from action $a_{f}$. The features $x'_u, u=\{1,\ldots,d\} \setminus v$ remain unobserved to the agent. 
%We can then simplify the observation model as $p(o \mid s', a) := \delta(o=x'_v)$. Note that while the state transition may be stochastic, the observation model is actually deterministic. 
The reward function $\mathcal{R}:\mathcal{S} \times \mathcal{A}_{c} \rightarrow \mathbb{R}$ specifies the reward structure for the original task MDP; the cost function $\mathcal{C}: \mathcal{S} \times \mathcal{A}_{f} \rightarrow \mathbb{R}_{\geq 0}$ defines the cost of acquisition actions. $\gamma \in [0,1]$ is the discount factor.

Given the partial observation history and the action history, we let $b$ represent a belief distribution over possible states. When the agent takes an action $a$ based on its policy $\pi(a \mid b)$, it receives the immediate reward $r = \mathcal{R}(s, a) \mathbb{I}(a \in \mathcal{A}_{c}) - \mathcal{C}(s,a) \mathbb{I}(a \in \mathcal{A}_{f})$. Our goal is to learn a policy $\pi(a \mid b)$ that maximizes the expected cumulative discounted reward $\mathbb{E}_{\pi,\mathcal{T}} \left[ \sum_{h=1}^{H} \gamma^h r \right]$, where $H$ represents the horizon of an episode. Next, we describe two types of the AA-POMDP that acquires features in different manners.

\textbf{Batch AA-POMDP}\quad
In the above AA-POMDP formulation, the action space consists of two types of actions, i.e., the feature acquisition actions $a_f \in \mathcal{A}_{f}$ and the task control actions $a_c \in \mathcal{A}_{c}$. We can accordingly decompose the policy $\pi(a \mid b)$ into two sub-policies:
$\pi_{f}$, which controls what acquisition actions to perform, and $\pi_{c}$, which controls what task-level actions to perform,
%to deal with the two types of actions respectively, i.e.,
\begin{equation}\label{eq:batch_acq}
    \pi(a \mid b) = \pi_{f}(a_{f} \mid b) \pi_{c}(a_{c} \mid b'),
\end{equation}
where the belief $b'$ is updated after acquiring the features indicated by the acquisition action $a_{f}$.
This formulation implies that the features are acquired simultaneously in a batch. For the $d$-dimensional feature space, the acquisition action space $\mathcal{A}_{f}=2^{[d]}$ is the powerset of all features, where $[d]$ represents the set $\{1,\ldots,d\}$. Each action $a_{f} \in \mathcal{A}_{f}$ indicates the subset of features being acquired. That is, the acquired features are $x_v, v=a_{f} \subseteq \{1,\ldots,d\}$.

The batch acquisition paradigm is useful when features are so time-critical that all acquisitions need to be performed in parallel to save time. For example, in an emergency, the doctor might need to acquire certain features as soon as possible to decide on a first aid strategy.
Another situation where the batch acquisition may help is when the acquisition action can modify the underlying state or the state may change between two acquisitions, such as the invasive procedures.
% For example, many imaging centers combine PET scans with CT or MRI to make sure the images are co-registered.

\textbf{Sequential AA-POMDP}\quad
In addition to batch acquisition, we also introduce the sequential acquisition scheme, where the features are acquired one-by-one in a sequence. Each acquisition action will acquire one of the features $\{1,\ldots,d\}$. We also introduce a special acquisition action $\phi$ to indicate the termination of the acquisition process. Therefore, the acquisition action space becomes $\mathcal{A}_{f} = \{1,\ldots,d\} \cup \{\phi\}$. Given the sequential acquisition process, we can further decompose \eqref{eq:batch_acq} to
\begin{equation}
    \pi(a \mid b) = \pi_{f}(a_{f} \mid b) \pi_{c}(a_{c} \mid b^\prime) =  \prod_{k=1}^{K} \pi_{f}(a_{f}^{(k)} \mid b^\prime_{k-1}) \pi_{c}(a_{c} \mid b^\prime_{K}),
\end{equation}
where 
$b^\prime_{k}$ is the belief after $k$ acquisition steps ($b^\prime_{0}=b$), 
%$K-1$
$K$ 
is the number of acquired features, and the last acquisition action is always $\phi$. Note that the beliefs $b^\prime_{k}$ are updated based on all the previously acquired features;
%, including the ones acquired with actions $a_f^{(<k)}$. 
the updated belief represents a more accurate distribution of the underlying state, and thus enabling better acquisition plan.

The sequential acquisition scheme may further reduce redundancy due to the awareness of the values from previous acquisitions, but at the expense of increased acquisition time because the acquisitions are performed sequentially. Therefore, this formulation is only applicable when the acquisition action is fast relative to the state changing. 
% For example, the MRI of a patient can be thought stable within some time period. 
Another implication of this formulation is that the acquisition action will not change the underlying state, otherwise, the previous observations will be outdated when making the task decisions.

\section{Methods}
In this section, we first develop a generative model to capture the dependencies of features along the state transition trajectory. The sequential generative model is leveraged afterwards to impute the missing features, which represent the belief of the agent. We then construct the feature acquisition policy and the task control policy in a hierarchical way based on the belief estimation.

\subsection{Partially Observed Sequence Modeling}\label{sec:poseq}
%For an episode of length $H$, l
Let $t \in \{1, \ldots, T\}$ denote 
% the time steps 
a time step
where state transition happens due to the execution of task action $a_c^{(t)}$ in the environment. At each time step $t$, our agent has acquired a subset of features $x_v^{(t)}$ from underlying state $s^{(t)}$, and the features $x_u^{(t)}$ remain unobserved to the agent.
% For an episode of length $T$, at each time step $t \in [1,T]$, our agent acquires a subset of features $x_v^{(t)}$ based on the acquisition action $a_f^{(t)}$ and executes a task action $a_c^{(t)}$ in the environment. The features $x_u^{(t)}$ remain unobserved to the agent. Afterwards, the environment transitions to a new state and the agent performs the acquisition and task repeatedly until the task is completed. 
In order to model the state transitions and estimate the beliefs about the underlying state, we build a generative model to impute the unobserved features conditioned on the observed ones and the action sequence: 
\begin{equation}\label{eq:poseq}
p(x_u^{(1:T)} \mid x_v^{(1:T)}, a_c^{(0:T-1)}).
\end{equation}
To simplify notation, we denote $a_c^{(0)}$ as a dummy action that initializes the environment. Note that the conditionals could be evaluated on an arbitrary subset of features since the agent may acquire different features for different instances. The conditionals essentially capture dependencies between the subset of features and across time steps.

\begin{figure}
    \centering
    \subfigure[Sequential Formulation]{\label{fig:poseq_seq}
    \includegraphics[width=0.26\linewidth]{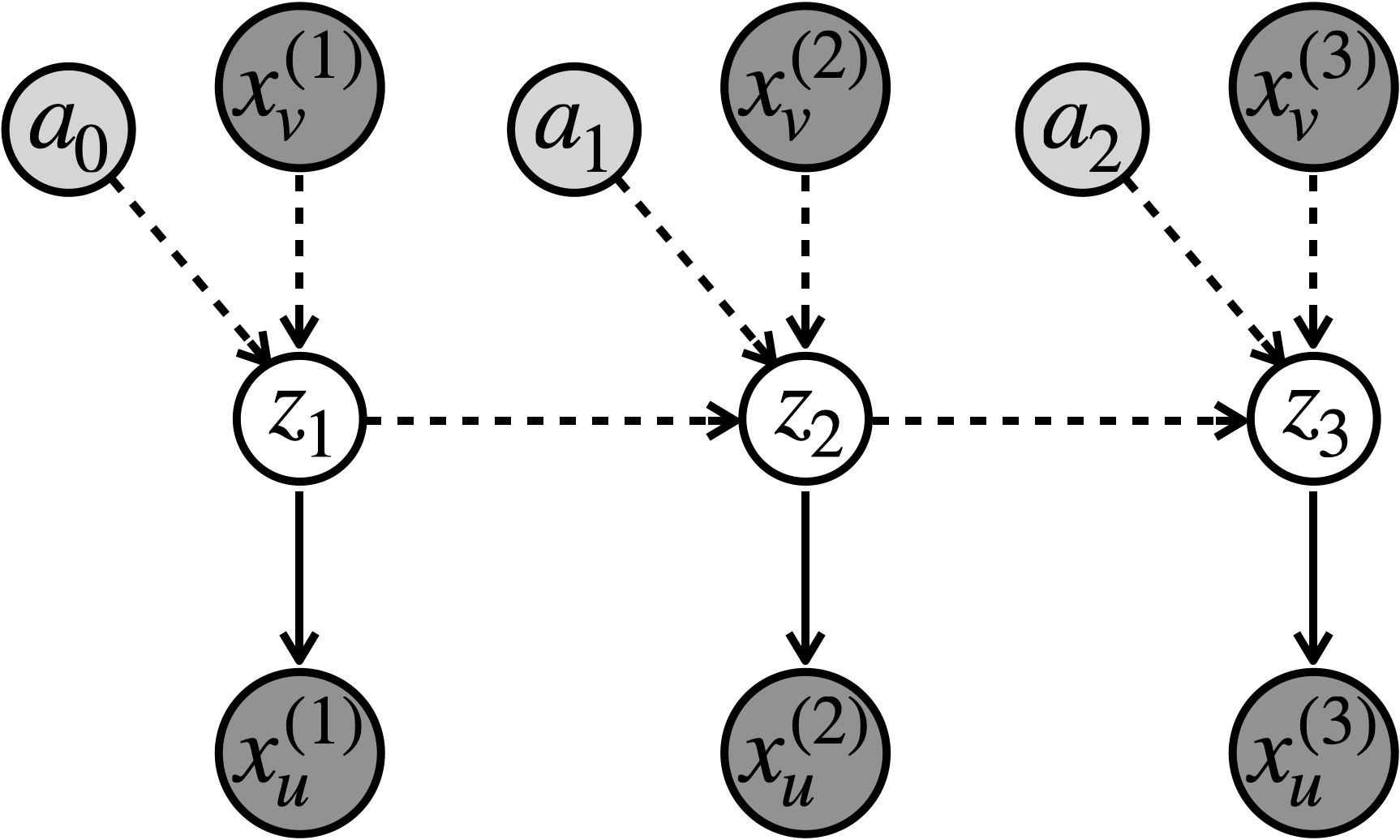}}
    \hspace{1cm}
    \subfigure[Set Formulation]{\label{fig:poseq_set}
    \includegraphics[width=0.34\linewidth]{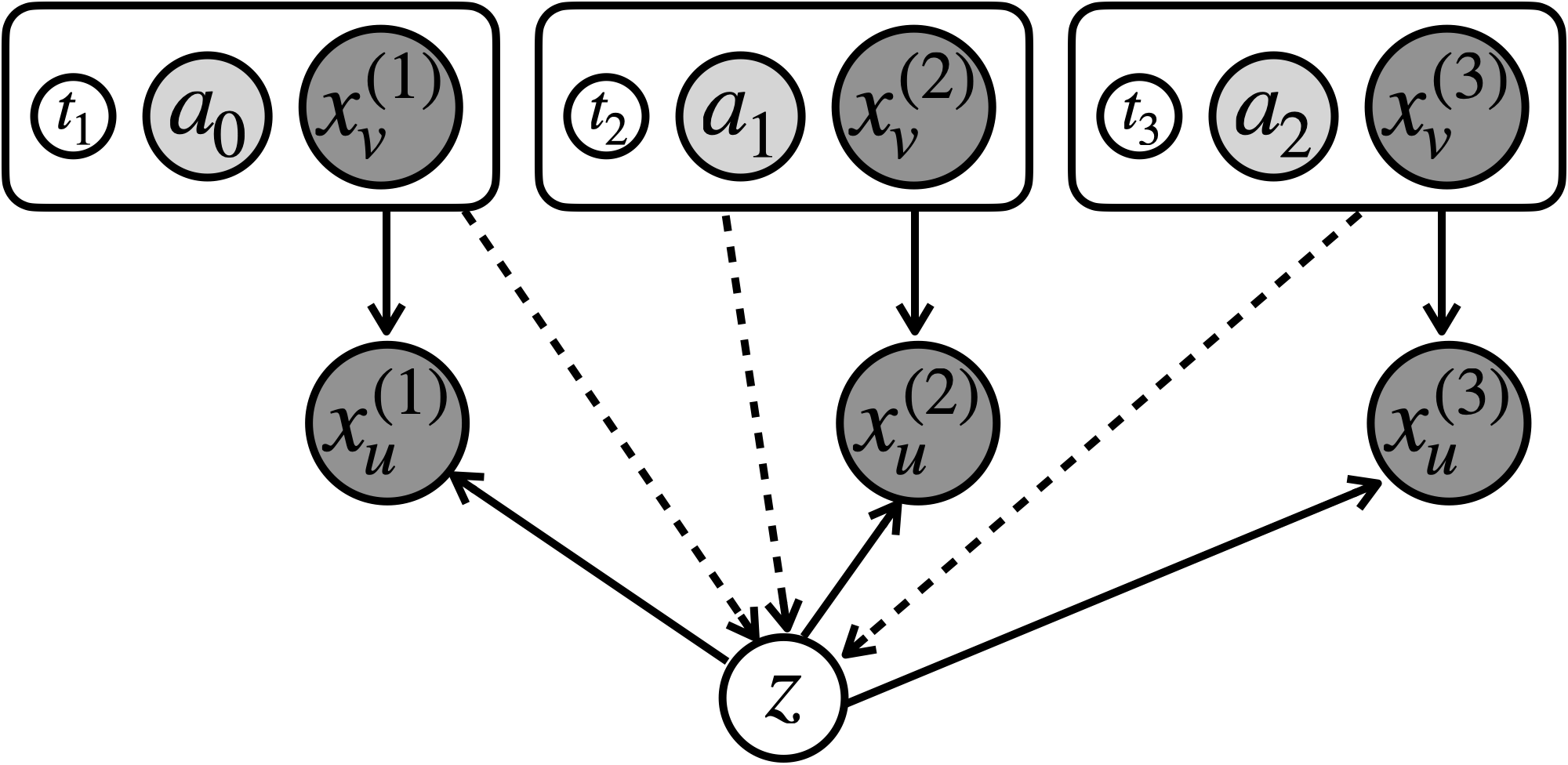}}
    \caption{Graphical model for modeling a trajectory with 3 time steps. The dashed arrows indicate the inference process, and the solid arrows indicate the generation process.}
    \label{fig:poseq}
\end{figure}

One way to model the conditional in \eqref{eq:poseq} is to exploit its sequential nature and factorize it by 
% \begin{equation}\label{eq:poseq_seq}
% \begin{aligned}
$
    p(x_u^{(1:T)} \mid x_v^{(1:T)}, a_c^{(0:T-1)}) = \prod_{t=1}^{T} p(x_u^{(t)} \mid x_u^{(1:t-1)}, x_v^{(1:t)},a_c^{(0:t-1)})
$
% \end{aligned}
% \end{equation}
A sequential latent variable $z^{(t)}$ can be introduced to simplify the model as in many sequential VAEs \cite{chung2015recurrent,igl2018deep,zhu2020s3vae,yin2020reinforcement}. Please see Fig.~\ref{fig:poseq_seq} for an illustration. However, the sequential formulation has several drawbacks: First, the latent variable is updated sequentially, which means the latent only depends on previous time steps, therefore the training signals coming from later time steps cannot be leveraged. Second, due to the limitation of recurrent models, the previous time steps might not have significant influence at the current time step, especially when the episode is long. Third, in order to make a prediction at a distant time step, the model has to unroll the latent multiple times, which could make the error accumulated and result in erroneous predictions.

In order to overcome those drawbacks, we draw inspiration from set modeling \cite{bender2020exchangeable,li2020exchangeable,li2021partially,kim2021setvae,bilovs2021scalable} and Transformer \cite{shan2021nrtsi,fang2021transformer,petrovich2021action} literature and formulate our generative modeling task in \eqref{eq:poseq} as a conditional set generation problem. Specifically, we concatenate the time index with the corresponding features and actions as a tuple and then the sequence becomes a permutation invariant set $\{(t, x_v^{(t)}, x_u^{(t)}, a^{(t-1)})\}_{t=1}^{T}$. We can then reformulate \eqref{eq:poseq} as
\begin{equation}\label{eq:poseq_set}
    p(\{x_u^{(t)}\}_{t=1}^{T} \mid \{(t, x_v^{(t)}, a_c^{(t-1)})\}_{t=1}^{T}) \equiv p(\mathbf{x}_u \mid \mathbf{ax}_v),
\end{equation}
where we denote $\mathbf{x}_u := \{x_u^{(t)}\}_{t=1}^{T}$ and $\mathbf{ax}_v := \{(t, x_v^{(t)}, a_c^{(t-1)})\}_{t=1}^{T}$ for notation simplicity.
Our Partially Observed Set models for Sequences (POSS) precisely overcomes the shortcomings of the aforementioned sequential generative models.
Based on the set formulation, we can now draw samples at arbitrary time points without having to rolling out the sequence step-by-step. During training, later time steps can propagate gradients to early ones and even distant time points can influence each other. Please see Fig.~\ref{fig:poseq_set} for an illustration.

To learn the conditional distribution over sets, we employ De Finetti's theorem \cite{diaconis1980finite,kerns2006definetti,edwards2016towards,li2021partially} and introduce a latent variable $z$. Given the latent variable, the conditionals can be decomposed:
%independently over time steps, i.e.,
\begin{equation}\label{eq:poseq_ind}
    p(\mathbf{x}_u \mid \mathbf{ax}_v) = \int \prod_{t=1}^{T} p(x_u^{(t)} \mid z, t, x_v^{(t)}, a_c^{(t-1)}) p(z \mid \mathbf{ax}_v) dz.
\end{equation}
However, optimizing \eqref{eq:poseq_ind} is still intractable due to the high dimensional integration over $z$. Therefore, we propose to utilize a variational approximation and optimize a lower bound:
\begin{equation}\label{eq:poseq_elbo}
    \log p(\mathbf{x}_u \mid \mathbf{ax}_v) \geq \sum_{t=1}^{T} \mathbb{E}_{q(z \mid \mathbf{ax})}\log p(x_u^{(t)} \mid z, t, x_v^{(t)}, a_c^{(t-1)}) -\KL(q(z \mid \mathbf{ax}) \| p(z \mid \mathbf{ax}_v)),
\end{equation}
where $\mathbf{ax}$ denotes the set $\{(t,x_u^{(t)},x_v^{(t)},a_c^{(t-1)})\}_{t=1}^{T}$,
which includes all of the features in $x^{(t)}$. 
$q(z \mid \mathbf{ax})$ and $p(z \mid \mathbf{ax}_v)$ are variational posterior and prior respectively, and they are permutation invariant w.r.t. the conditioning set inputs. $p(x_u^{(t)} \mid z, t, x_v^{(t)}, a_c^{(t-1)})$ is the decoder distribution, which could be shared for each time step $t$. Note that different from typical VAE models, the decoder operates over arbitrary subset of features. That is, the decoder takes in a subset of observed features along with other conditional information and outputs the distribution for the remaining subset of unobserved features. 
Please see Sec.~\ref{sec:appendix_poss} in the appendix for detailed derivations and model illustration.
% Please see Sec.~\ref{sec:appendix_poss} in the appendix for detailed derivations and see Fig.~\ref{fig:poseq_elbo} for an illustration of the model.

To deal with the arbitrary dimensionality for feature subsets $x_u^{(t)}$ and $x_v^{(t)}$, we impute the missing features with zeros and introduce a binary mask to indicate whether the corresponding dimensions are observed or not. That is, for the prior and the decoder, $x_v^{(t)}$ is represented as the concatenation of a $d$-dimensional features and a $d$-dimensional binary mask, where the missing features are replaced by zeros; while for the posterior, we combine $x_u^{(t)}$ and $x_v^{(t)}$ and regard all features as observed.

Given the complexity of set based inputs and arbitrary dimensionality of observed features, modeling the posterior and prior using a simple distribution family, such as Gaussian, may not be optimal. Therefore, we propose to use normalizing flows for both prior and posterior distributions. Following the best practice in normalizing flow literature \cite{kingma2016improved,durkan2019neural}, we model the posterior using inverse autoregressive flow for its fast sampling speed. The prior is modeled using a coupling flow with spline networks. The base distribution for both distributions are Gaussian conditioned on their corresponding set representations. However, the ELBO in \eqref{eq:poseq_elbo} is now not analytically available due to normalizing flow based posterior and prior distributions. We instead using a Monte Carlo estimation by sampling multiple (M) latent $z_m$ from the posterior:
\begin{equation}\label{eq:poseq_nflow}
    \frac{1}{M} \sum_{t=1}^{T} \sum_{m=1}^{M} \left[ \log p(x_u^{(t)} \mid z_m, t, x_v^{(t)}, a_c^{(t-1)}) - \log q(z_m \mid \mathbf{ax}) + \log p(z_m \mid \mathbf{ax}_v) \right].
\end{equation}
During training, we assume access to both the observed features $x_v^{(t)}$ and the unobserved features $x_u^{(t)}$. Therefore, we can directly optimize the ELBO in \eqref{eq:poseq_nflow}. During sampling, given a set of observed features $\{x_v^{(t)}\}_{t=1}^{T}$ and the corresponding actions $\{a_c^{(t-1)}\}_{t=1}^{T}$, we can impute the unobserved features at any time steps, even at time steps beyond $T$.

\subsection{Belief State Estimation}
In order to solve the aforementioned AA-POMDP, our agent will need to determine an optimal acquisition plan and an optimal task action sequence based solely on the partially observed information. Fortunately, the sequential generative model can impute the missing features and thus estimate the belief about the underlying state.

At any specific time step $h$, suppose the agent has executed task actions $a_c^{(<h)} \equiv \{a_c^{(i-1)}\}_{i=1}^{h}$ resulting in the underlying state $s^{(h)}$, the agent has access to the observation history $o^{(<h)} \equiv \{x_v^{(i)}\}_{i=1}^{h-1}$ \footnote{At time $i-1$, the agent might have taken a feature acquisition action, so $a_c^{(i-1)}$ might be undefined. For notation simplicity, here $a_c^{(i-1)}$ actually means the last task action the agent have taken before time $i-1$.}.
%When beginning the hierarchical policy, t
The acquisition sub-policy will begin with $x_v^{(h)}=\varnothing$.
In the sequential setting $x_v^{(h)}$ shall be updated with each acquisition sub-step (with the acquired feature values from state $s^{(h)}$); 
in the batch acquisition setting, the observation $x_v^{(h)}$ is updated only once after all specified acquisitions are made.  
Given the available information, we utilize  the sequential generative POSS model ($\S$\ref{sec:poseq}) to predict the unobserved features for state $s^{(h)}$, i.e., $x_u^{(h)}$. We first sample a latent code from the prior $p(z \mid \{(i,x_v^{(i)}, a_c^{(i-1)})\}_{i=1}^{h})$, then pass the latent code through the decoder only for state $s^{(h)}$ to obtain the distribution $p(x_u^{(h)} \mid z, t, x_v^{(h)}, a_c^{(h-1)})$, to sample the unobserved features $x_u^{(h)}$.
% At any specific time step $h$, suppose the agent has executed task actions $a_c^{(<h)} \equiv \{a_c^{(i-1)}\}_{i=1}^{t}$ resulting in the underlying state $s^{(t)}$, the agent has access to the observation history $o^{(<h)} \equiv \{x_v^{(i)}\}_{i=1}^{t}$. For batch acquisition scenario, the observation $x_v^{(t)}$ may be empty if the action $a^{(h)}$ at time step $h$ is supposed to be an acquisition action. For sequential acquisition scenario, $x_v^{(t)}$ may contain the immediately acquired feature values from state $s^{(t)}$. 
% Given the available information, we resort to the sequential generative model to predict the unobserved features for state $s^{(t)}$, i.e., $x_u^{(t)}$. We first sample a latent code from the prior following $p(z \mid \{(i,x_v^{(i)}, a_c^{(i-1)})\}_{i=1}^{t})$, then passing the latent code through the decoder only for state $s^{(t)}$ to obtain the distribution $p(x_u^{(t)} \mid z, t, x_v^{(t)}, a_c^{(t-1)})$, from which the unobserved features $x_u^{(t)}$ can be sampled.

The distribution over the unobserved features may have multiple modes, therefore, using one sample may not accurately represent the beliefs. We instead perform multiple imputations by sampling multiple latent codes. The belief at time step $h$ can then be represented as a set of imputed features, i.e., $b^{(h)} = \{(x_v^{(h)}, \hat{x}_u^{(h)})_n\}_{n=1}^{N}$, where $N$ is the number of samples of the unobserved features.

\subsection{Cost Sensitive Reinforcement Learning}\label{sec:csrl}
Given the sequential transition model and the belief estimates, we now build the RL agent to solve the AA-POMDP. We decompose the agent into two policies, the feature acquisition policy $\pi_f$ and the task policy $\pi_c$, which are then combined in a hierarchical fashion. Both policies take the belief estimation set $b^{(h)}$ as inputs and derive the action distribution in a permutation invariant manner.

\begin{figure}
    \centering
    \subfigure[Batch Acquisition]{
    \includegraphics[width=0.32\linewidth]{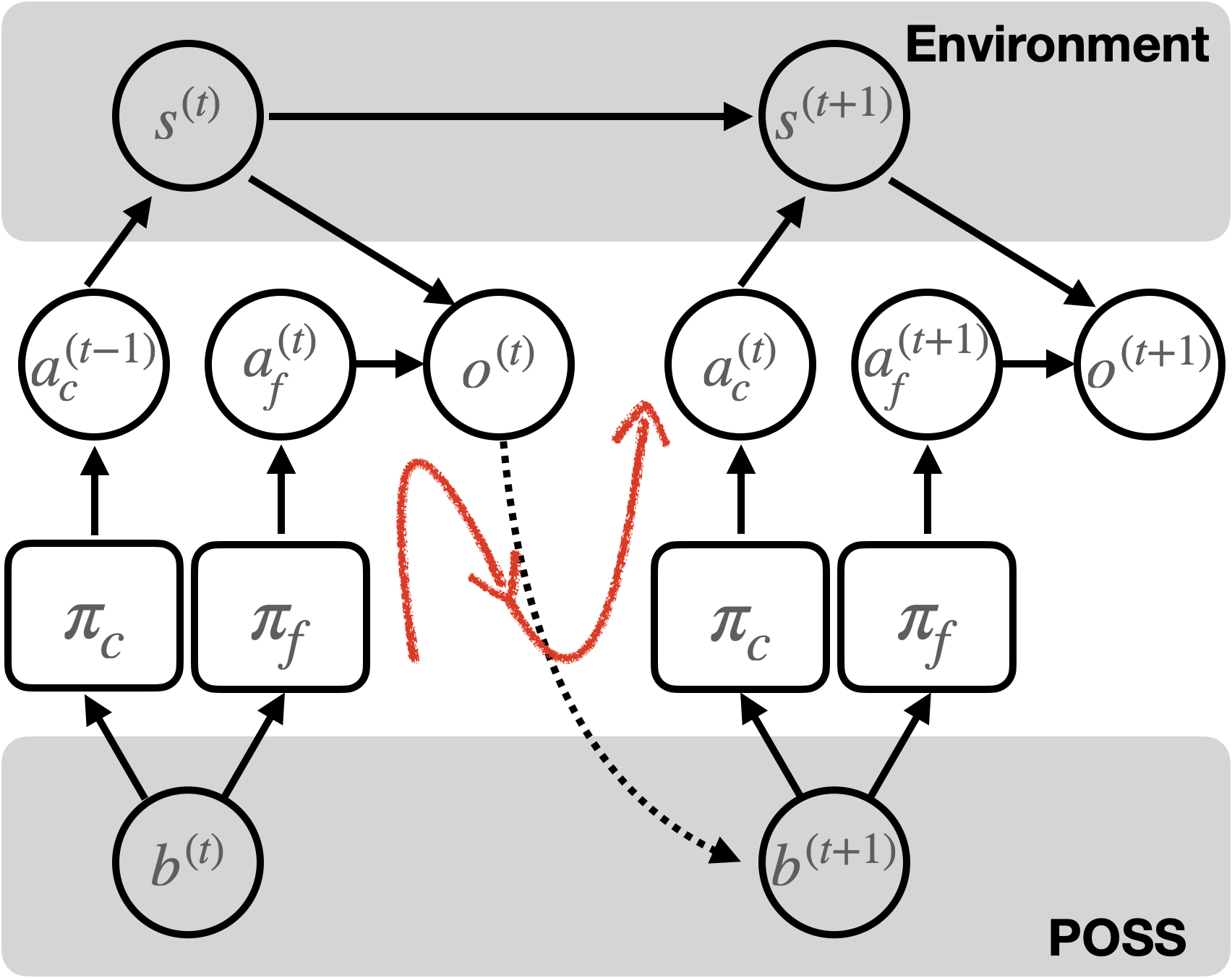}}
    \hspace{1cm}
    \subfigure[Sequential Acquisition]{
    \includegraphics[width=0.35\linewidth]{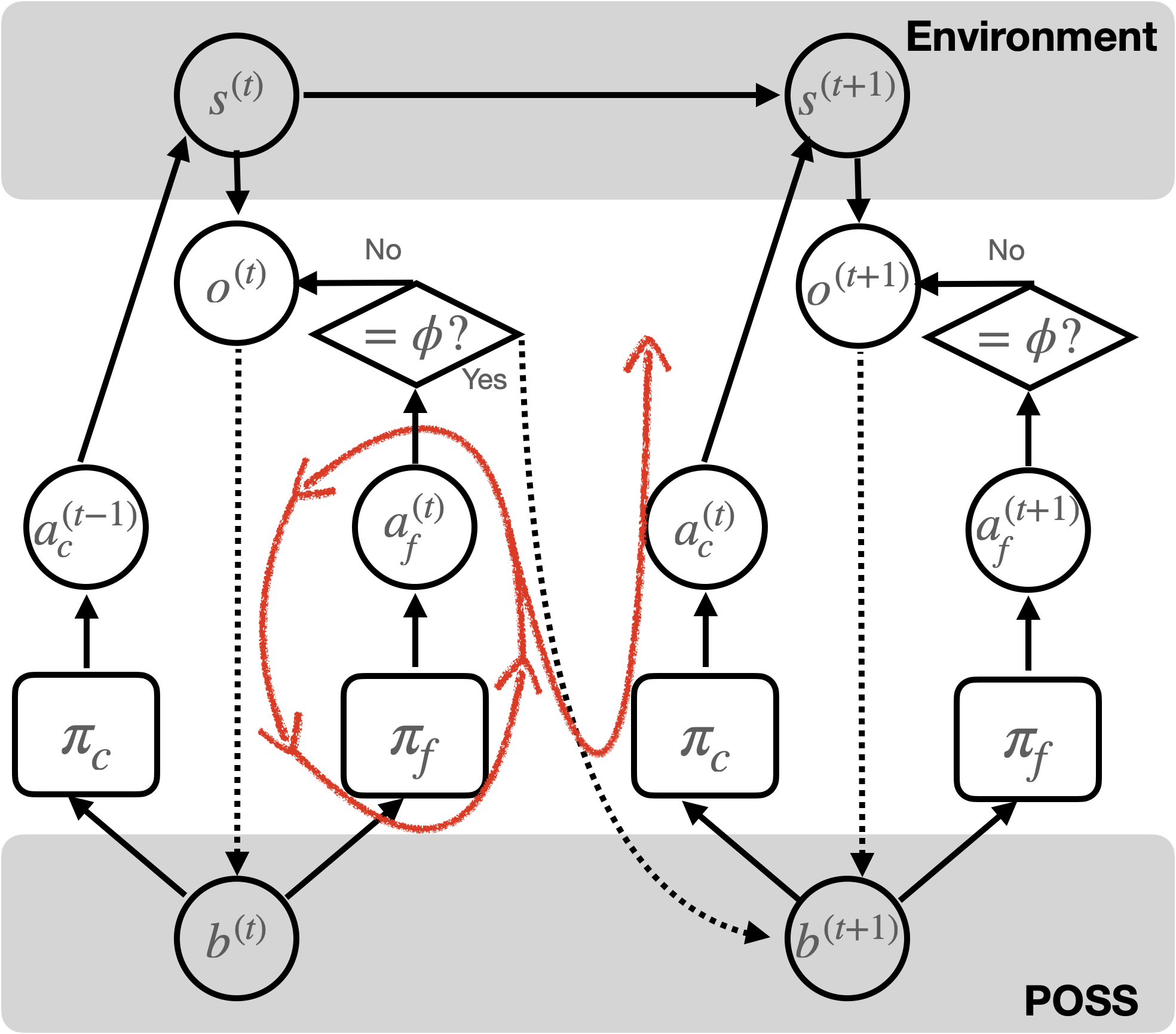}}
    \caption{Illustrations of the batch acquisition process and the sequential acquisition process. The dashed lines indicate the update of the belief. The red arrows represent the hierarchical policy execution processes.}
    \label{fig:csrl}
\end{figure}

At any underlying state $s^{(h)}$, we first run the feature acquisition policy $\pi_f$ to collect information. The features are acquired either in a batch or one-by-one depending on the acquisition setting. In the batch acquisition setting, the acquisition policy is ran once to determine the set of features to be acquired, while in the sequential acquisition setting, the acquisition policy is run multiple times sequentially. The belief estimations are updated after acquiring the features. The task policy $\pi_c$ is then executed based on the updated belief using the acquired features (see Fig.~\ref{fig:csrl} for illustrations). %See Fig.~\ref{fig:csrl} for illustrations of our hierarchical policies.

Our goal in the AA-POMDP is to maximize the task reward while minimizing the feature acquisition cost. In the hierarchical setting, we decompose the goal as well. The high-level task policy aims at achieving as high task reward as possible based on the acquired features, while the low-level acquisition policy aims at providing sufficient information and minimizing the acquisition cost. 
Therefore, the reward for the task policy at time step $h$ is defined as 
%\begin{equation}
$r_c^{(h)} = \mathcal{R}(s^{(h)}, a_c^{(h)})$,
%\end{equation}
which is the same as the original task reward.
For the acquisition policy, in addition to the acquisition cost, we desire the acquired features to support the task policy. First, the task policy should produce confident action choice based on the acquired features. Therefore, we use the negative entropy of the task policy as a reward to the acquisition policy, i.e., $-\text{Ent}(\pi_c(b^{(h)}))$. Second, the acquired features as inputs to the task policy should lead to a high value estimation indicating high long-term return of the task policy. Therefore, we employ the task value estimation, $V_c(b^{(h)})$, as an additional reward. Third, the acquired features should be indicative of the unobserved features so that the belief estimation is accurate. We hence use the imputation accuracy as one of the acquisition rewards, i.e., $\text{Acc}(x_u^{(h)}, \hat{x}_u^{(h)})$. For discrete features, the accuracy is evaluated as the average exact mach accuracy of the $N$ belief samples. For continuous features, the accuracy is evaluated as the average negative MSE of the $N$ belief samples. In total, the reward for the acquisition policy at time step $h$ is defined as
\begin{equation}
    r_f^{(h)} = -\mathcal{C}(s^{(h)}, a_f^{(h)}) - \omega_e \cdot \text{Ent}(\pi_c(b^{(h)})) + \omega_v \cdot V_c(b^{(h)}) + \omega_a \cdot \text{Acc}(x_u^{(h)}, \hat{x}_u^{(h)}),
\end{equation}
where $\omega_e$, $\omega_v$ and $\omega_a$ are hyperparameters for weighting the corresponding terms. In the batch acquisition setting, all features in $x_v^{(h)}$ are acquired simultaneously, thus the rewards are received immediately after the acquisition. In the sequential acquisition setting, however, each acquisition action will only receive its cost as immediate reward, while the other reward terms are granted only when the agent selects the termination action $\phi$.

\subsection{Implementation}
In this section, we describe several important implementation details. We use PPO \cite{schulman2017proximal} for both the acquisition policy and the task policy. The actor and the critic networks are implemented as an ensemble over the belief sets, where the action probabilities and values are averaged over the belief set elements. The sequential generative model is implemented as a VAE with normalizing flow based posterior and prior, of which the base distribution are Gaussian conditioned on the corresponding sets. We use Set Transformers \cite{lee19d} to extract the set representations. The time indexes are embedded using sinusoidal functions as in other Transformer models \cite{vaswani2017attention}. For continuous actions and features, we directly concatenate them. For discrete actions and features, we learn their embeddings jointly. During training, we first train the sequential generative model with trajectories obtained by randomly acquired features and random task actions; then we pre-train the task policy with fixed generative model and random acquisitions; finally we train the generative model and both policies jointly.

\section{Related Works}

\textbf{Active Feature Acquisition and Active Perception} \quad
Active feature acquisition is a relevant field where features are actively acquired with costs to predict the target variable. Previous works \cite{zubek2002pruning,ruckstiess2011sequential,shim2018joint,yoon2019asac,chang2019dynamic} have formulated the AFA problem as MDP and have developed various of reinforcement learning approaches to find the optimal acquisition plan for a given instance. \cite{li2021active} further propose a model-based solution by leveraging ACFlow \cite{li2019flow} to model the AFA dynamics. The learned dynamics then assists the agent by providing auxiliary information and intrinsic rewards. \cite{zannone2019odin} propose to learn the acquisition policy using augmented data sampled from a pretrained Partial VAE \cite{ma2018eddi}. \cite{he2012imitation} and \cite{he2016active} instead employ the imitation learning approach guided by a greedy reference policy to learn the acquisition policy. 
In addition to RL based approaches, \cite{ling2004decision}, \cite{chai2004test} and \cite{nan2014fast} propose decision tree, naive Bayes, and maximum margin based classifiers, respectively, to jointly minimize the misclassification cost and feature acquisition cost. \cite{ma2018eddi}, \cite{gong2019icebreaker} and \cite{li2020dynamic} propose to acquire features greedily using mutual information as the estimated utility. 
In contrast to the existing AFA works, our setting does not contain a specific target variable; instead, we focus on optimizing the cumulative reward of MDP. Furthermore, our agent learns the feature acquisition policy and the task policy simultaneously.
Active perception is a relevant sub-field where a robot with a mounted camera is planning by selecting the best next view \cite{bajcsy1988active,aloimonos1988active, cheng2018reinforcement,jayaraman2018learning}. 

% Active perception is a relevant sub-field where a robot with a mounted camera is planning by selecting the best next view \cite{bajcsy1988active,aloimonos1988active, cheng2018reinforcement,jayaraman2018learning}. Successful applications include object recognition \cite{karayev2012timely,ammirato2017dataset,jayaraman2016look,johns2016pairwise,malmir2015deep}, object detection \cite{caicedo2015active,garcia2015active,mathe2016reinforcement}, robot navigation \cite{gupta2017cognitive,mirowski2016learning,zhu2017target}, etc. Here, instead of observing different views of the environment, our agent will observe a subset of features from the underlying state. 

\textbf{POMDP and Temporal Dynamics Modeling}\quad
Learning in POMDP without access to the environment model is much more difficult than learning in MDP \cite{papadimitriou1987complexity}. Many works, therefore, have focused on planning in POMDP with a known environment model \cite{littman1995learning,hauskrecht2000value,pineau2003point,ross2007aems,ross2008online,kurniawati2008sarsop,silver2010monte,somani2013despot,bai2014integrated,sunberg2018online}. Bayes-Adaptive POMDP \cite{ross2007bayes,ross2011bayesian,katt2018bayesian} instead learn the environment model in a Bayesian fashion by assuming access to an informative prior over the observation model and plan using posterior belief distributions over states. Instead of planing with an environment model, Deep Recurrent Q-Networks (DRQN) \cite{hausknecht2015deep} and its variants \cite{zhu2017improving} parameterize the value function with a recurrent neural network that takes in the action and observation history. The value function is later learned using the DQN RL algorithm \cite{mnih2015human}. Deep Variational Reinforcement Learning (DVRL) \cite{igl2018deep} uses the action and observations history to learns a VAE model, where the latent variable is interpreted as the belief. The A2C RL algorithm \cite{wu2017scalable} is then applied on the latent representation and trained together with the generative model. TD-VAE \cite{gregor2018temporal} builds a VAE model to predict the belief state for time points separated by random intervals. Their jumpy state modeling enables the prediction of belief at arbitrary future time without the step-by-step rollout.

Outside of POMDP literature, there is a number of works that consider jumps when modeling temporal dynamics. \cite{koutnik2014clockwork} and \cite{chung2016hierarchical} equip recurrent neural network with skip connections, which makes it easier to bridge distant time steps. \cite{buesing2018learning} temporally sub-sample the data with fixed jump interval and build models on the subsampled data. One of the limitations of the subsampling is that the model cannot leverage information contained in the skipped observations. \cite{neitz2018adaptive} and \cite{jayaraman2018time} predict sequences with variable time-skips, by choosing the most predictable future frames as target.

Due to the feature acquisition, our setting makes the agent observe only a subset of features, thus following a POMDP, but with the difference that the observation is controlled by the agent itself rather than the environment. In order to infer the belief and assist policy learning, we develop a VAE model to impute the missing features. In our model, the action and observation history together with their timestamps are treated as a permutation invariant set. The set perspective enables our model to directly predict the belief at arbitrary time step without resorting to the stepwise rollout. 

\textbf{Cost Sensitive Reinforcement Learning}\quad
Previous works have considered learning agent to decide \emph{what} and \emph{when} to observe when the observation has a cost. 
\cite{zubek2000pomdp} introduce a special type of POMDP called even-odd POMDP, in which the world is assumed to be fully observable every other time step. They then convert it into an equivalent MDP, whose value function captures the sensing costs of the original POMDP. \cite{zubek2004two} propose the Cost Observable MDP (COMDP), where the actions are partitioned into those that change the state of the world and those that are pure observation actions, the reward function has been modified to incorporate observation costs. The COMDP is conceptually similar to our AA-POMDP, but their algorithm focuses solely on tabular environments and the batch acquisition scenario.  \cite{yin2020reinforcement} study the batch acquisition scenario in continuous-state POMDP. A sequential VAE model is trained offline to extract representation form the action and observation history, and then an RL policy takes in the latent representation and outputs both acquisition action and task action. \cite{bellinger2020active} propose to learn a policy and a state estimator in parallel during online training. The agent either pays the cost to observe the full state or trust the estimated state for free. \cite{nam2021reinforcement} propose Action-Contingent Noiselessly Observable MDPs (ACNO-MDPs), a special class of POMDPs in which the agent either fully observe the state at a cost or act without any immediate observation, relying on past observations to infer the underlying state. \cite{bellinger2021scientific} consider a similar intermittently observed scenario and provide a in-depth qualitative analysis of agents’ measurement patterns for two RL algorithms, Dueling DQN \cite{wang2016dueling} and PPO \cite{schulman2017proximal}. In this work, we study both the batch acquisition and sequential acquisition scenarios. Our agent can observe an arbitrary subset of features at each decision step, which is a generalization of the ACNO-MDPs studied in \cite{bellinger2020active,nam2021reinforcement,bellinger2021scientific,Krale2023ActThenMeasureRL}.

\section{Experiments}\label{sec:exp}
We evaluate batch acquisition and sequential acquisition scenarios on several benchmark environments. 
For context, we provide the rewards stemming from a task policy on \texttt{fully\_observed} states.
Additionally, we also tested a typical POMDP setting where a random subset of features were observed (\texttt{random*}).
We vary the cost per acquisition to demonstrate the trade-off between acquisition cost and task reward. We evaluate both the batch acquisition setting and sequential acquisition setting with our proposed cost-sensitive hierarchical PPO (CS-HPPO) ($\S$~\ref{sec:csrl}), where the belief state is estimated by POSS ($\S$~\ref{sec:poseq}). In order to verify the benefits of our belief estimation, we compare to the variants
that replace belief with the observation history, which is a typical practice in POMDP literature \cite{McCallum1993OvercomingIP}. 
\begin{wrapfigure}{r}{0.3\linewidth}
    \centering
    \begin{minipage}{0.98\linewidth}
        \subfigure[batch]{
            \includegraphics[width=0.98\linewidth]{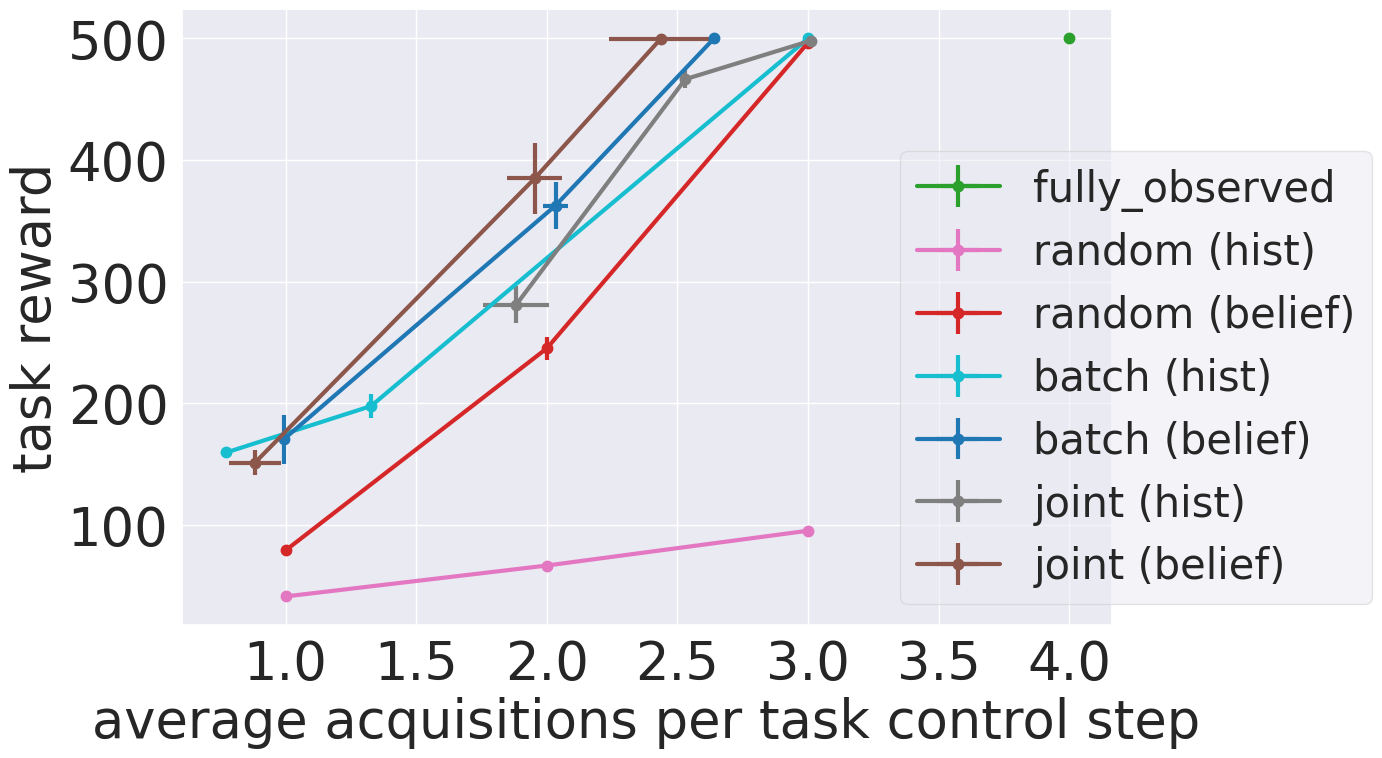}
        }
        \subfigure[sequential]{
            \includegraphics[width=0.98\linewidth]{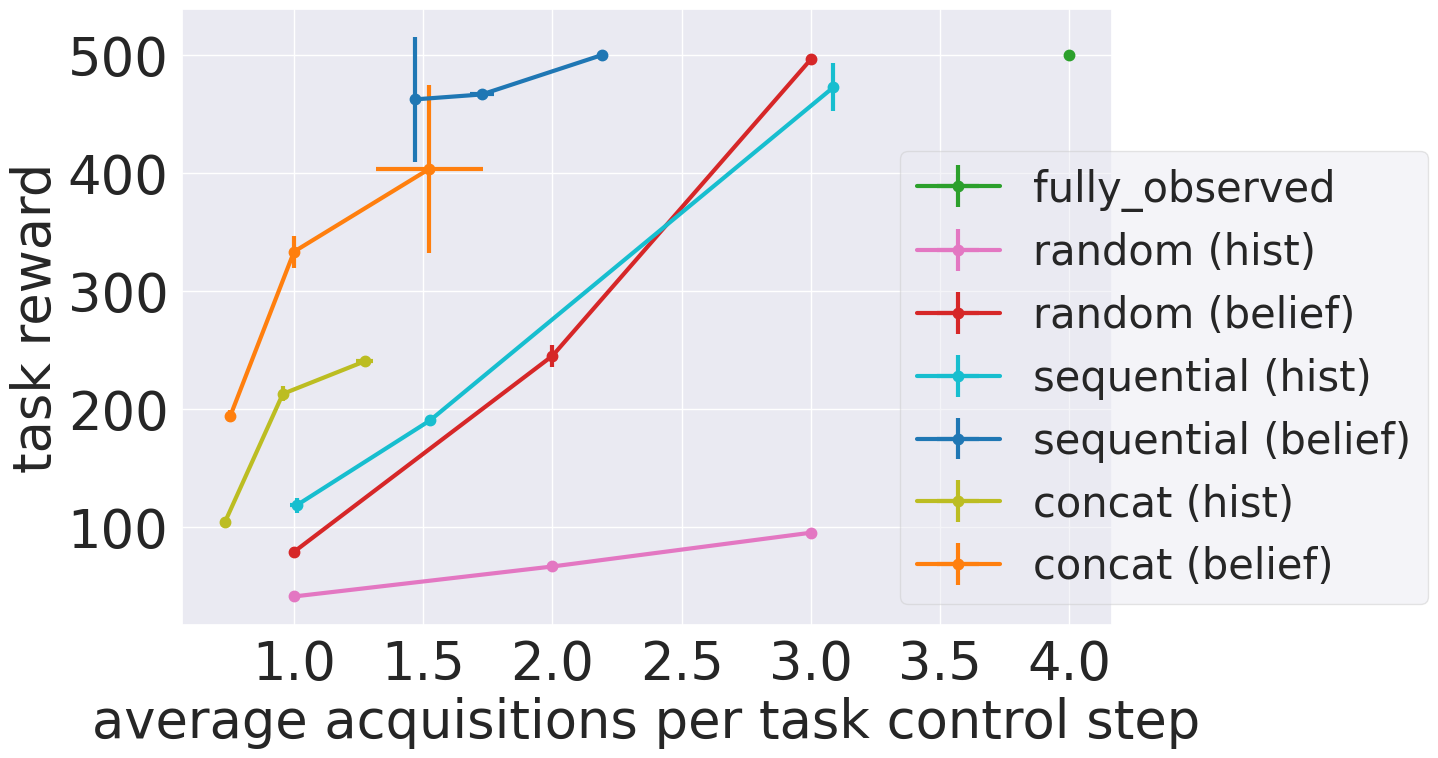}
        }
        \caption{CartPole results}
        %\vspace{-20pt}
        \label{fig:cartpole_results}
    \end{minipage}
\end{wrapfigure}
I.e., PPO agents select an action at each step based on either the belief estimation (\texttt{belief}) or the observation history (\texttt{hist}).
Inspired by \cite{nam2021reinforcement}, we compare our batch acquisition models to a setting
where the action space is the Cartesian product of task control actions and feature acquisition actions (\texttt{joint*}). In this setting, the acquisition action controls what will be observed in next state.
%instead of the current state, and the agent observes all features at the beginning. 
For sequential acquisition setting, we also compare to a setting that concatenates feature acquisition actions and task control actions into a larger action space (\texttt{concat*}).
%then a PPO agent selects an action at each step based on either the belief estimation or the observation history.  
We attempted to evaluate a 
generic POMDP-RL algorithm, DRQN \cite{hausknecht2015deep}, in the setting of concatenated action space as well, but found that it fails to learn effective acquisition policy. Please find more details in Appendix~\ref{sec:appendix_exp}. 
%For comparison, we also provide the results with a random feature acquisition policy and rewards in the fully observed setting. 

\textbf{Partially Observed CartPole}\quad
First, we evaluate on a modified OpenAI gym CartPole-v1 environment, where the features of a state can be dynamically acquired with a cost. In the batch acquisition setting, the action space contains 16 acquisition actions and 2 task control actions, while in the sequential acquisition setting, the acquisition action spaces contains the four measurable features plus a termination action. 

\textbf{Sepsis Simulator}\quad
This environment simulates a Sepsis patient and the effects of several common treatments \cite{oberst2019counterfactual}. The task is to apply three treatment actions, antibiotic, ventilation and vasopressors, to the ICU patients. Therefore, the task action space is the powerset of the three
\begin{wrapfigure}{r}{0.3\linewidth}
    \centering
    \begin{minipage}{0.98\linewidth}
        \centering
        \subfigure[batch]{
            \includegraphics[width=0.98\linewidth]{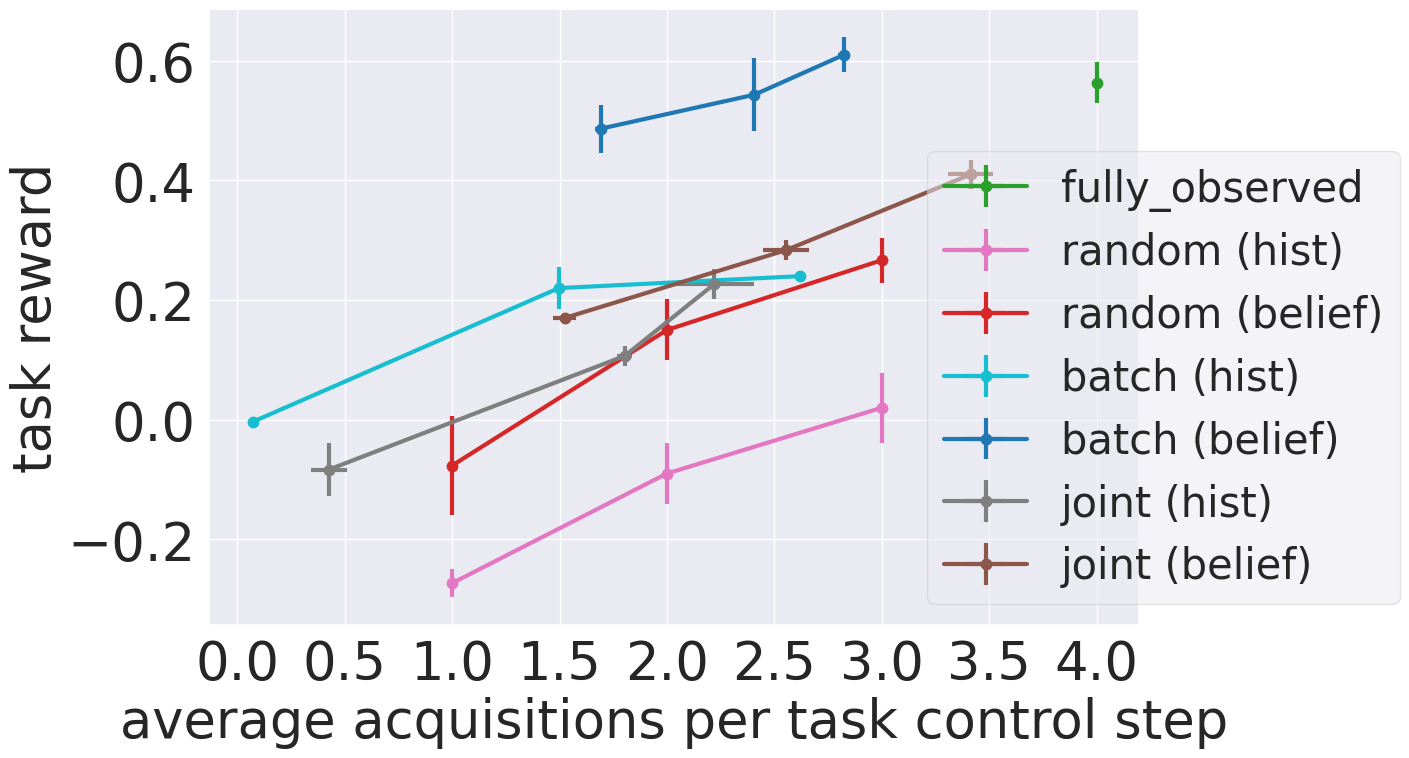}
        }
        \subfigure[sequential]{
            \includegraphics[width=0.98\linewidth]{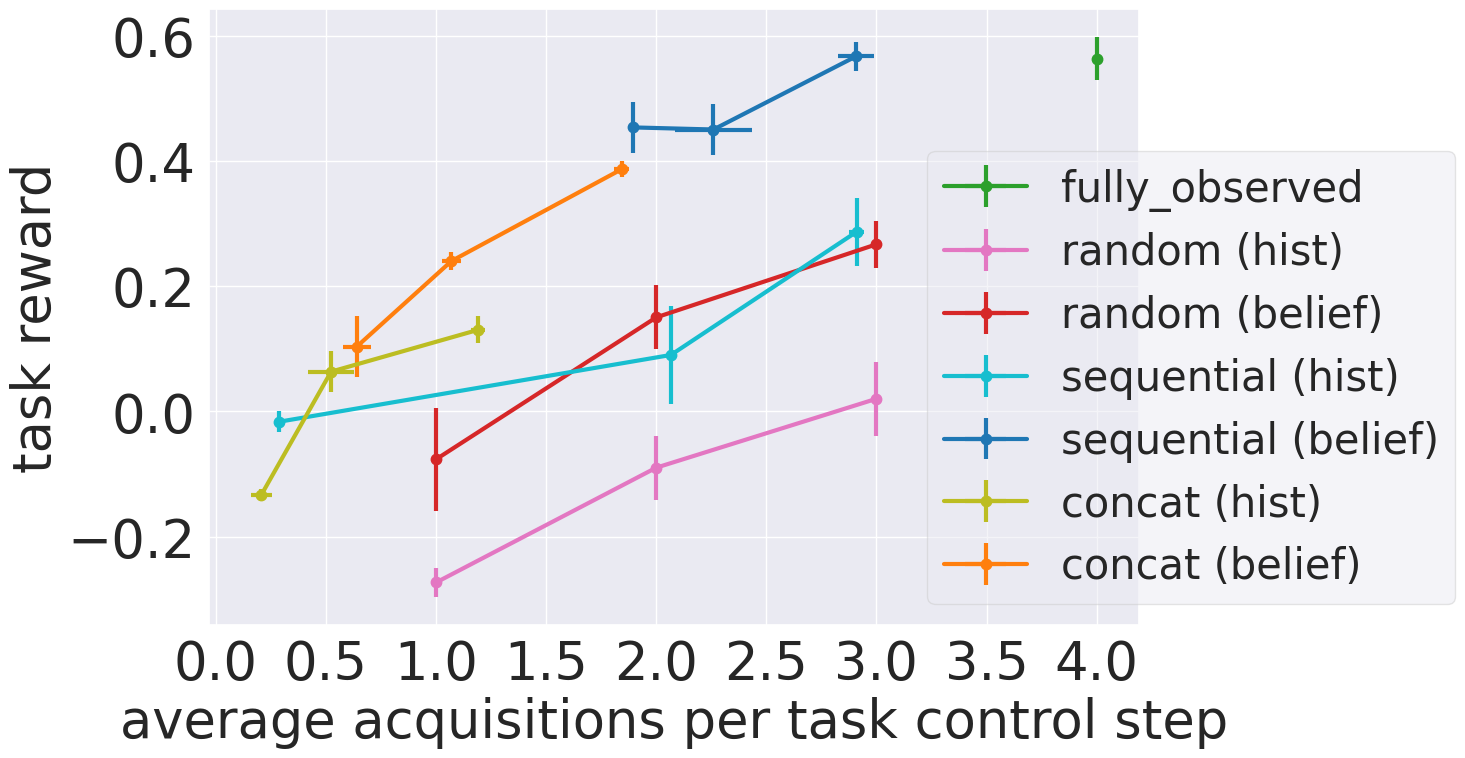}
        }
        \caption{Sepsis results.}
        \vspace{10pt}
        \label{fig:sepsis_results}
    \end{minipage}
    
    \begin{minipage}{0.98\linewidth}
            \centering
            \includegraphics[width=.9\linewidth]{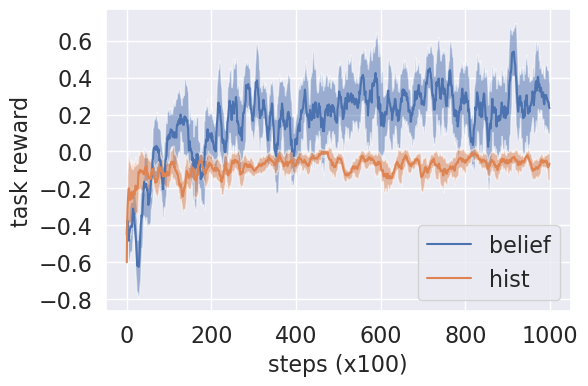}
            \caption{Training curve.}
            \label{fig:training_curve}
    \end{minipage}
\end{wrapfigure}
treatments. Each patient has eight features including one indicator of the patent’s diabetes condition, three indicators of 
the current treatment state and four measurable states over heart rate, sysBP rate, 
percoxyg state and glucose level. We only allow the agent to acquire the four measurable states and regard the 
rest of features as given. Therefore, in the batch acquisition setting, the acquisition action space 
contains the powerset of the four measurable features, i.e., 16 acquisition actions in total; in the sequential acquisition setting, the acquisition action space contains the four measurable features plus a termination action. We limit the maximum treatment steps to 30. The patient will be discharged if
his/her measurement states all return to normal values, which gives the agent a $1.0$ reward. An episode will also terminate upon mortality with a reward $-1.0$.

% \begin{figure}
%     \begin{minipage}{0.49\linewidth}
%         \subfigure[batch]{
%             \includegraphics[width=0.46\linewidth]{results/cartpole_batch_small.png}
%         }
%         \subfigure[sequential]{
%             \includegraphics[width=0.46\linewidth]{results/cartpole_seque_small.png}
%         }
%         \caption{CartPole results (best viewed digitally).}
%         \label{fig:cartpole_results}
%     \end{minipage}
%     \hfill
%     \begin{minipage}{0.49\linewidth}
%         \centering
%         \subfigure[batch]{
%             \includegraphics[width=0.46\linewidth]{results/sepsis_batch_small.png}
%         }
%         \subfigure[sequential]{
%             \includegraphics[width=0.46\linewidth]{results/sepsis_seque_small.png}
%         }
%         \caption{Sepsis results (best viewed digitally).}
%         \label{fig:sepsis_results}
%     \end{minipage}
% \vspace{-10pt}
% \end{figure}

\textbf{Results}\quad
Figure~\ref{fig:cartpole_results} and \ref{fig:sepsis_results} demonstrate the results for CartPole and Sepsis respectively. We run each acquisition setting with three acquisition costs and three random seeds. For each acquisition cost, we plot the average task reward for the original control task against the average number of acquired features for each task control action. We can see that in every acquisition setting (either batch acquisition, sequential acquisition, random acquisition, acquisition in concatenated action space, or acquisition in joint action space), the agent equipped with the 
belief estimation performs much better than the ones using observation history. During training, we also observed the belief estimation help stabilize the training and converges to the optimal policy quickly. Please see Fig.~\ref{fig:training_curve} for an example of the training curve. For agents that have access to the 
belief estimation, our proposed CS-HPPO almost always outperforms the non-hierarchical policies in both batch acquisition and sequential acquisition settings. One exception is the batch acquisition setting in CartPole environment, where acquisition in the joint action space performs better. We 
believe it is due to the relatively small action space (only 32 actions in the joint action space) so that the non-hierarchical policy is good enough to explore the space while the hierarchical one introduces additional complexity. 

% \begin{wrapfigure}{r}{0.25\linewidth}
%     \centering
%     \includegraphics[width=\linewidth]{results/log.png}
%     \caption{Example training curve.}
%     \label{fig:training_curve}
% \end{wrapfigure}
\begin{wraptable}{r}{0.28\linewidth}
    \tiny
    \centering
    \begin{tabular}{c|cc}
    \toprule
         & ID & OOD \\
    \midrule
       Seq-PO-VAE  & 93.32 & 75.12 \\
       POSS & \textbf{94.49} & \textbf{78.54} \\
    \bottomrule
    \end{tabular}
    \caption{Prediction accuracy for in-distribution (ID) and oud-of-distribution (OOD) time steps.}
    \label{tab:poss}
\end{wraptable}

\textbf{Ablation Studies}\quad
As an ablation study, we compare the prediction accuracy of our proposed sequential generative model (POSS) to the Seq-PO-VAE proposed in \cite{yin2020reinforcement}. For a fair comparison, we augment Seq-PO-VAE with normalizing flow based prior and posterior distributions as in POSS. We train both models on 1000 trajectories collected from Sepsis simulator with random acquisition policy and random task policy and test on the held-out 100 trajectories. 
To evaluate the generalizability, we train models using only observations from the first 15 trajectory steps and test the prediction accuracy on both in-distribution (ID) time steps (<=15) and out-of-distribution (OOD) time steps (>15).
%To evaluate the generalizability of POSS, we train the model using only observations from the first 15 time steps of each trajectory and test the prediction accuracy on both in-distribution time steps (<=15) and out-of-distribution time steps (>15). 
Table~\ref{tab:poss} shows accuracies and we can see POSS outperforms Seq-PO-VAE on both ID and OOD time steps, verifying the superiority of our proposed set based sequence modeling approach.

\section{Conclusion}
In this work, we study the sequential decision making problems with feature acquisition costs. We present a special MDP named AA-POMDP and identify two types of the feature acquisition settings, batch acquisition and sequential acquisition, which are applicable under different conditions. To help solve the partially observed problem, we develop a sequential generative model to capture the state transitions multiple imputation of the unobserved features. The agent then takes a set of imputed observations as the belief estimation. In order to balance the acquisition cost with the task reward, we propose a hierarchical formulation of the policy, where the low-level policy is responsible for acquiring features and the high-level policy maximizes the task reward based on the acquired feature subsets. The entire framework, including both the generative model and two levels of the policies, is trained jointly. We conduct extensive experiments and demonstrate state-of-the-art performance.

\bibliographystyle{unsrt}
\bibliography{neurips_2023}

\clearpage
\appendix

\counterwithin{figure}{section}
\renewcommand\thefigure{\thesection.\arabic{figure}}
\counterwithin{table}{section}
\renewcommand\thetable{\thesection.\arabic{table}}
\counterwithin{equation}{section}
\renewcommand\theequation{\thesection.\arabic{equation}}

\section{Partially Observed Set Models for Sequences (POSS)}\label{sec:appendix_poss}
As discussed in Sec.~\ref{sec:poseq}, we formulate the partially observed sequence modeling task \eqref{eq:poseq} as a set modeling task \eqref{eq:poseq_set} for a set $\mathbf{ax} := \{(t, x_v^{(t)}, x_u^{(t)}, a_c^{(t-1)})\}_{t=1}^{T}$. According to De Finetti's theorem, there exists a latent code $z$ such that the set elements are conditionally independent conditioned on $z$, i.e.,
\begin{equation}
\begin{aligned}
    p(\mathbf{ax}) &= p(\{(t, x_v^{(t)}, x_u^{(t)}, a_c^{(t-1)})\}_{t=1}^{T})\\
    &= \int \prod_{t=1}^{T} \left[ p(t, x_v^{(t)}, x_u^{(t)}, a_c^{(t-1)} \mid z) \right] p(z) dz \\
    &= \int \prod_{t=1}^{T} \left[ p(x_u^{(t)} \mid t, x_v^{(t)}, a_c^{(t-1)}, z) p(t, x_v^{(t)}, a_c^{(t-1)} \mid z) \right] p(z) dz \\
    &= \int \prod_{t=1}^{T} \left[ p(x_u^{(t)} \mid t, x_v^{(t)}, a_c^{(t-1)}, z) \right] \prod_{t=1}^{T} \left[ p(t, x_v^{(t)}, a_c^{(t-1)} \mid z) \right] p(z) dz \\
    &\stackrel{(1)}{=} \int \prod_{t=1}^{T} \left[ p(x_u^{(t)} \mid t, x_v^{(t)}, a_c^{(t-1)}, z) \right] p(\{(t, x_v^{(t)}, a_c^{(t-1)})\}_{t=1}^{T} \mid z) p(z) dz\\
    &= \int \prod_{t=1}^{T} \left[ p(x_u^{(t)} \mid t, x_v^{(t)}, a_c^{(t-1)}, z) \right] p(z \mid \{(t, x_v^{(t)}, a_c^{(t-1)})\}_{t=1}^{T}) p(\{(t, x_v^{(t)}, a_c^{(t-1)})\}_{t=1}^{T}) dz \\
    &\equiv \int \prod_{t=1}^{T} \left[ p(x_u^{(t)} \mid t, x_v^{(t)}, a_c^{(t-1)}, z) \right] p(z \mid \mathbf{ax}_v) p(\mathbf{ax}_v) dz.
\end{aligned}
\end{equation}
The equation $(1)$ applies the De Finetti's theorem again. Since $x_v^{(t)}$ contains a subset of features at time step $t$, the same latent variable $z$ that factors the set element $(t, x_v^{(t)}, x_u^{(t)}, a_c^{(t-1)})$ conditionally independent will also factor $(t, x_v^{(t)}, a_c^{(t-1)})$ conditionally independent.

Divide both sides with $p(\mathbf{ax}_v) := p(\{(t, x_v^{(t)}, a_c^{(t-1)})\}_{t=1}^{T})$, we have
\begin{equation}\label{eq:appendix_poss}
    p(\mathbf{x}_u \mid \mathbf{ax}_v) = \int \prod_{t=1}^{T} \left[ p(x_u^{(t)} \mid t, x_v^{(t)}, a_c^{(t-1)}, z) \right] p(z \mid \mathbf{ax}_v) dz.
\end{equation}

\begin{figure}[h!]
    \centering
    \includegraphics[width=0.95\linewidth]{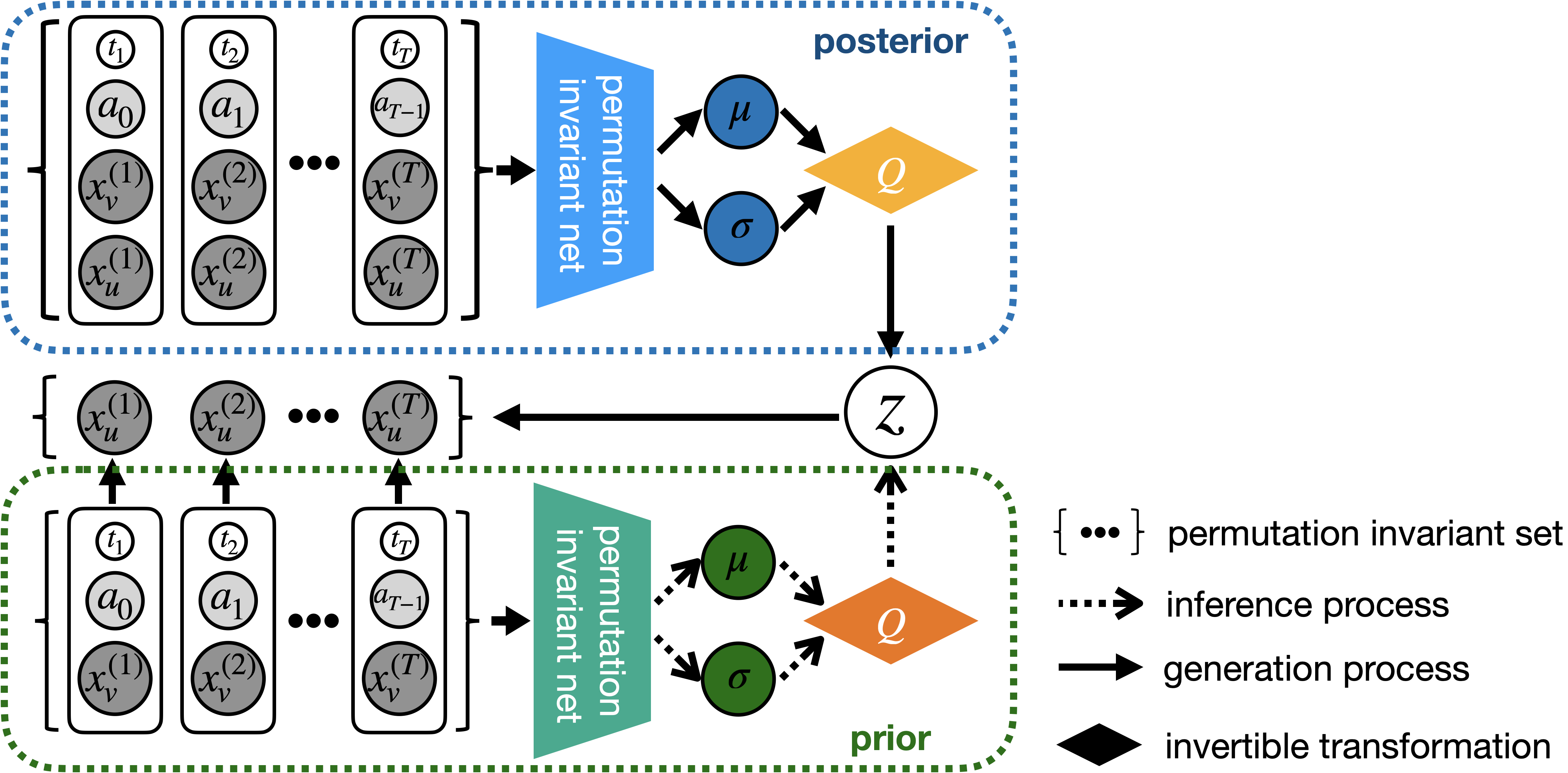}
    \caption{Model partially observed trajectories via set based VAE.}
    \label{fig:poseq_elbo}
\end{figure}

To optimize \ref{eq:appendix_poss}, we resort to the variational approach and optimize a lower bound \eqref{eq:poseq_elbo}. The prior $p(z \mid \mathbf{ax}_v)$ and posterior $q(z \mid \mathbf{ax})$ are permutation invariant w.r.t. their inputs $\mathbf{ax}_v$ and $\mathbf{ax}$ respectively. To obtain accurate estimations of the prior and posterior, we utilize normalizing flow based distributions where the base distributions are parameterized as Gaussian distributions with mean and variance derived from $\mathbf{ax}_v$ and $\mathbf{ax}$ using Set Transformers. Due to the permutation invariant architecture of Set transformer, the Gaussian base distribution is permutation invariant; and since the transformations are invertible, the ultimate normalizing flow based distributions are permutation invariant as well. Please see Fig.~\ref{fig:poseq_elbo} for an illustration of our proposed POSS model.

% \section{Cost Sensitive Reinforcement Learning}\label{sec:appendix_csrl}

\section{Experiment}\label{sec:appendix_exp}
In this section, we evaluate both the batch acquisition setting and the sequential acquisition setting with our proposed cost-sensitive hierarchical PPO (CS-HPPO). The following are a list of settings we experimented:

\paragraph{Fully Observed} In this setting, the agent only needs the task policy $\pi_c$ since all features will be observed at each time step. The task policy takes the full observation as input and no belief estimation is needed either.

\paragraph{Random Acquisition} Since there will not be an acquisition policy, we cannot control the number of acquisitions using cost. Instead, we set a fixed budget so that the agent can observe part of the features that is selected at random. We evaluate two variants of this setting where the task policy takes input from either the previous 8 observations or the belief estimated by POSS. 

\paragraph{Batch Acquisition} When the agent arrives at state $s^{(h)}$, the agent first runs the acquisition policy $\pi_f$ to select a set of features to acquire; then, based on the acquired features, the agent runs the task policy $\pi_c$ to perform the actual task. If the agent is equipped with POSS, the belief is updated right after each acquisition so that the next task action is based on the updated belief; otherwise, the agent takes the previous 8 observations as input for both acquisition policy and task policy.

\paragraph{Joint Action Space} Acquisition in joint action space is meant to be a baseline for batch acquisition, where we combine the batch acquisition actions and task actions by their Cartesian product to form a joint action space. At the beginning, the agent observes all features and then selects a joint action where the task action transits the environment to a new state and the acquisition action determines what features to observe for next step. The following actions are selected based only on the acquired features. The agent continues this process until the task is terminated. We similarly evaluate two variants of this setting where the inputs to the policy are either the previous observations or the belief estimation.

\paragraph{Sequential Acquisition} In the sequential acquisition setting, the agent first runs the acquisition policy to acquire features from the underlying state $s^{(h)}$ and terminates acquisition until it selects the termination action $\phi$. Afterwards, the task policy $\pi_c$ selects a task action based on the acquired features. The inputs to the policies are either belief estimations or observation histories.

\paragraph{Concatenated Action Space} As a baseline for sequential acquisition, we evaluate a setting where the acquisition actions and task acquisitions are concatenated. At each step, the agent selects either an acquisition action or a task action. When the agent selects the task action, it automatically terminates the acquisition process and transits the environment to a new state by executing the task action within the environment. Similarly, the inputs to the policy could be belief estimation or observation histories.
 
\subsection{Benchmark Environments}

\subsubsection{Partially Observed CartPole}
The OpenAI gym CartPole-v1 environment contains 4 features (i.e., cart position, cart velocity, pole angle and pole angular velocity) and 2 discrete actions (i.e., push cart to left and push cart to right). In the batch acquisition setting, the action space contains $2^4=16$ acquisition actions and 2 task control actions. In the sequential acquisition setting, the acquisition action space contains the 4 measurable features plus a termination action $\phi$, and the task action space contains the 2 original task control actions. We conduct experiments with three different costs per feature ($0.005$, $0.01$, and $0.015$), and for each cost we report results from 3 independent runs. Since each run might acquire different number of features and achieve different rewards, we report the mean and standard deviation for both the acquisitions per task action and the task reward.

\subsubsection{Sepsis Simulator}
As described in Sec.~\ref{sec:exp}, the Sepsis simulator contains 8 features, in which 4 of them can be acquired, while the rest 4 features are given. The agent can take 8 treatment actions. In the batch acquisition setting, the action space contains $2^4 = 16$ acquisition actions and 8 task actions. In the sequential acquisition setting, the acquisition action space contains the 4 measurable features plus a termination action $\phi$, and the task action space contains the 8 treatment actions. We conduct experiments with three different costs per acquisition ($0.005$, $0.01$, and $0.02$), and report results from 3 independent runs for each cost.

\subsection{POSS Implementation}
The POSS model contains a prior network, a posterior network, two invertible transformations for prior and posterior respectively and a decoder network. The prior and posterior networks first use 4 permutation equivariant Set Transformer layers with 128 hidden units to extract set based features; then, an attentive pooling layer squashes the set features into a 128 dimensional permutation invariant feature vector; finally, 2 linear layers with 128 hidden units output the mean and variance for the Gaussian base distribution. we set the latent variable dimension to 64. For prior, we stack 4 rational-quadratic coupling transformations to transform the base distribution; and for posterior, we use 4 rational-quadratic autoregressive transformations \cite{Durkan2019NeuralSF}. For categorical observations, we learn a set of 16 dimensional embeddings for each feature and a special embedding to represent the missing feature. We also embed the discrete actions with 16 dimensional features. The time steps are represented by sinusoidal functions as 16 dimensional features. The inputs for the prior network contain the time step embeddings, the action embeddings and the embeddings of observed features, while the inputs for posterior network contain the time step embedddings, the action embeddings and the embeddings of all features. The decoder network takes the latent code as well as time step embeddings, action embeddings and embeddings for observed features as inputs and outputs a distribution for the unobserved features. For categorical features, the decoder outputs logits of a Categorical distribution; and for continuous features, the decoder outputs mean and variance of a Gaussian distribution.

\subsection{Policies}
In different settings, the policy network will have different type of inputs. In fully observed setting, the policy takes in a vector representation of the observations. When using observation histories, the policy network takes in a set of partially observed features. When using belief estimations, the policy network takes in multiple imputations of the unobserved features. We use a 2-layer linear network to implement both the acquisition policy and the task policy. If the input is a set (history or belief), we obtain the final action distribution with ensemble. For discrete actions, the actor outputs a categorical distribution where the probabilities are the average probability across set elements. For continuous actions, the actor distribution is a Gaussian distribution where the mean is averaged across set elements. 

\subsection{Hyperparameters}
Table~\ref{tab:hyperparameters} list all the hyperparameters for CartPole and Sepsis environments. Note that we did not conduct any hyperparameter optimization and all hyperparameters are set based on our previous experiences.

\begin{table}[h!]
    \centering
    \begin{tabular}{cc|cc}
    \toprule
       Component & HyperParameter &  CartPole & Sepsis \\
    \midrule
       PPO & $\gamma$  &  0.99  & 0.99\\
       PPO & $\lambda$  & 0.95  & 0.95\\
       PPO & clip ratio & 0.2 & 0.2\\
       PPO & reward weight $\omega_e$ & 1.0 & 1.0\\
       PPO & reward weight $\omega_v$ & 0.01 & 1.0\\
       PPO & reward weight $\omega_a$ & 100.0 & 1.0\\
    \midrule
      Policy & actor & \multicolumn{2}{c}{Linear: $64 \rightarrow 64$}\\
      Policy & critic & \multicolumn{2}{c}{Linear: $64 \rightarrow 64$}\\
    \midrule
      POSS & prior & \multicolumn{2}{c}{SetTransformer: $128 \times 4$ $+$ Linear: $128 \rightarrow 64$}\\
      POSS & prior transformations & \multicolumn{2}{c}{rational-quadratic coupling: $128 \times4$}\\
      POSS & posterior & \multicolumn{2}{c}{SetTransformer: $128 \times 4$ $+$ Linear: $128 \rightarrow 64$}\\
      POSS & posterior transformations & \multicolumn{2}{c}{rational-quadratic autoregressive: $128 \times4$}\\
      POSS & decoder & \multicolumn{2}{c}{SetTransformer: $128 \times 4$ $+$ Linear: $128 \rightarrow 128$}\\
    \midrule
      Training & model learning rate & 0.0001 & 0.0001\\
      Training & actor learning rate & 0.0003 & 0.0003\\
      Training & critic learning rate & 0.0003 & 0.0003\\
      Training & grad norm & 1.0 & 1.0 \\
    \bottomrule
    \end{tabular}
    \caption{Hyperparameters}
    \label{tab:hyperparameters}
\end{table}

\subsection{Additional Ablation Studies}
The benefits of our proposed POSS are two folds: First, it provides the agent with an accurate belief estimation so that the agent can make better decisions based solely on the partial observations. Second, the imputation accuracy provides an intrinsic reward to the acquisition policy to guide the agent acquire informative features. We have seen the belief estimation help achieve better reward-cost tread-off compared to observation histories (Sec.~\ref{sec:exp}), and we have verified the advantage of the the set based sequence modeling formulation (Sec.~\ref{sec:exp}). To better understand the benefits of the intrinsic reward, we conduct an ablation study on Sepsis simulator by removing the intrinsic reward (set $\omega_a$ to 0). 

\begin{figure}[h!]
    \centering
    \subfigure[task reward]{
        \includegraphics[width=0.45\textwidth]{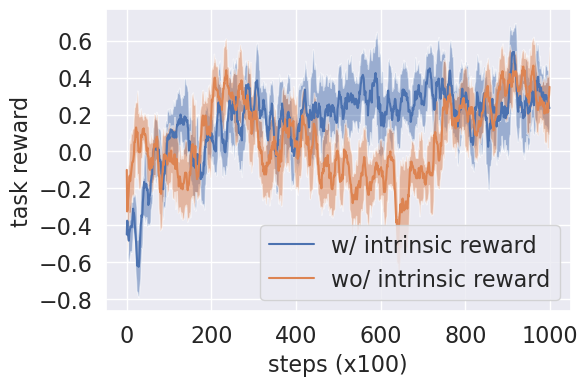}
    }
    \subfigure[acquisitions per task action]{
        \includegraphics[width=0.45\textwidth]{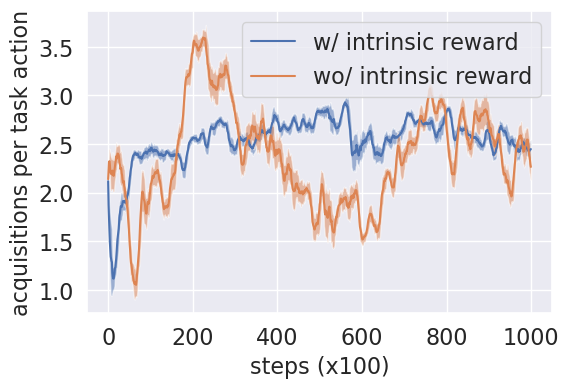}
    }
    \caption{Compare task reward (a) and average acquisitions per task action (b) for two agents with and without intrinsic reward provided by POSS.}
    \label{fig:ablation_reward}
\end{figure}

Figure~\ref{fig:ablation_reward} compares the training curve for the two agents with and without intrinsic reward. First, we can see the two agent eventually converges to a similar solution (both similar number of acquisitions and similar task reward), which empirically verifies that the intrinsic reward does not affect the optimal policy. Second, we can see the agent trained with intrinsic reward converges much faster and the training is more stable.

\end{document}